\documentclass{article}

      \usepackage[preprint]{neurips_2019}

\usepackage[utf8]{inputenc} 
\usepackage[T1]{fontenc}    
\usepackage[svgnames]{xcolor}
   \definecolor{codegreen}{rgb}{0,0.6,0}
   \definecolor{codegray}{rgb}{0.5,0.5,0.5}
   \definecolor{codepurple}{rgb}{0.58,0,0.82}
\usepackage{hyperref}       
   \hypersetup{colorlinks=true, breaklinks=true, linkcolor=DarkRed, urlcolor=DarkBlue, citecolor=DarkGreen}

\usepackage{url}            
\usepackage{booktabs}       
\usepackage{amsfonts}       
\usepackage{nicefrac}       
\usepackage{microtype}      
\usepackage{graphicx}
\usepackage{xspace}
\usepackage{amsmath}
\usepackage{amssymb}
\usepackage{multirow}
\usepackage{multicol}
\usepackage{adjustbox}
\usepackage{listings}
   \lstdefinestyle{mystyle}{
      frame=single,
      commentstyle=\color{codegreen},
      keywordstyle=\color{blue}\bfseries,
      numberstyle=\tiny\color{codegray},
      stringstyle=\color{codepurple},
      basicstyle=\linespread{0.68}\fontsize{6}{10.8}\ttfamily\bfseries,
      breakatwhitespace=false,         
      breaklines=false,                 
      captionpos=b,                    
      keepspaces=true,                 
      numbers=none,                    
      numbersep=3pt,                  
      showspaces=false,                
      showstringspaces=false,
      showtabs=false,                  
      tabsize=2,
      language=[Sharp]C,
      moredelim=**[is][\underbar]{_}{_},
      escapechar=\@
   }
\usepackage{subcaption} 
\usepackage{tikz}
\usepackage{flushend}
\usepackage{textcomp}
\usepackage{floatrow}
   \newfloatcommand{capbtabbox}{table}[][\FBwidth]
\usepackage{xxcolor}
\usepackage{tcolorbox}
   \tcbset{on line,size=small,boxsep=0.2em,right skip=0.2em}
\usepackage{todonotes}
\usepackage{tabularx}
\usepackage{wrapfig}
\usepackage{enumitem}
   \setlist{nosep,labelsep=*,leftmargin=*}

\setcitestyle{numbers,,open={[},close={]}}

\newcommand{\eg}{\hbox{\emph{e.g.}}\xspace}
\newcommand{\ie}{\hbox{\emph{i.e.}}\xspace}

\newcommand{\etc}{\hbox{\emph{etc.}}\xspace}
\newcommand{\resp}{\hbox{\emph{resp.}}\xspace}

\makeatletter
\newenvironment{btHighlight}[1][]
{\begingroup\tikzset{bt@Highlight@par/.style={#1}}\begin{lrbox}{\@tempboxa}}
  {\end{lrbox}\bt@HL@box[bt@Highlight@par]{\@tempboxa}\endgroup}
\newcommand\btHL[1][]{%
  \begin{btHighlight}[#1]\bgroup\aftergroup\bt@HL@endenv%
  }
  \def\bt@HL@endenv{%
  \end{btHighlight}%
  \egroup
}
\newcommand{\bt@HL@box}[2][]{%
  \tikz[#1]{%
    \pgfpathrectangle{\pgfpoint{1pt}{0pt}}{\pgfpoint{\wd #2}{\ht #2}}%
    \pgfusepath{use as bounding box}%
    \node[anchor=base west, fill=orange!25,outer sep=.5pt,inner xsep=0.5pt, inner ysep=-0.3pt, rounded corners=2pt, minimum height=\ht\strutbox-.1pt,#1]{\raisebox{.01pt}{\strut}\strut\usebox{#2}};
  }%
}
\makeatother

\newcommand{\dypro}{\textsc{DyPro}\xspace}
\newcommand{\coset}{\textsc{CoSet}\xspace}

\title{\coset: A Benchmark for Evaluating \\ Neural Program Embeddings}

\author{%
  Ke Wang \\
Visa Research \\
Palo Alto, CA 94306 \\
\href{mailto:kewang@visa.com}{kewang@visa.com} \\
\And
Mihai Christodorescu \\
Visa Research\\
Palo Alto, CA 94306 \\
\href{mailto:mihai.christodorescu@visa.com}{mihai.christodorescu@visa.com} \\
}

\begin{document}

\maketitle


\begin{abstract}
   Neural program embedding can be helpful in analyzing large software, a task that is challenging for traditional logic-based program analyses due to their limited scalability. A key focus of recent machine-learning advances in this area is on modeling program semantics instead of just syntax. Unfortunately evaluating such advances is not obvious, as program semantics does not lend itself to straightforward metrics.

   In this paper, we introduce a benchmarking framework called \coset for standardizing the evaluation of neural program embeddings. \coset consists of a diverse dataset of programs in source-code format, labeled by human experts according to a number of program properties of interest. A point of novelty is a suite of program transformations included in \coset. These transformations when applied to the base dataset can simulate natural changes to program code due to optimization and refactoring and can serve as a ``debugging'' tool for classification mistakes. We conducted a pilot study on four prominent models---TreeLSTM~\cite{TreeLSTM}, gated graph neural network (GGNN)~\cite{allamanis2017learning}, AST--Path neural network (APNN)~\cite{alon2018code2vec}, and \dypro~\cite{wang2017dynamic}. We found that \coset is useful in identifying the strengths and limitations of each model and in pinpointing specific syntactic and semantic characteristics of programs that pose challenges.
\end{abstract}


\section{Introduction}
\label{sec:intro}



Analyzing large software artifacts (so-called ``Big Code'') has proven extremely challenging for logic-base approaches, which quickly force a trade-off between accuracy and scalability. Applying deep learning architectures to this task is a promising direction that flourished in the last few years, with a number of techniques being proposed around a variety of features extracted from software programs, ranging from syntactic to statically derived from source code to dynamically inferred from execution traces. Since these models operate in separate application domains, it is hard to draw a comparison. Our goal in this paper is to measure how precisely models can capture the program semantics and to this end we propose a framework, \coset, to standardize the evaluation of neural program embeddings.

The idea is to reuse the knowledge distilled from existing code repositories in an attempt to simplify the future development of software, and in turn improve the product quality, is quite powerful. Code completion for example is a significant boost to programmer productivity and as such is a worthwhile task to pursue. Early methods in the field applied NLP techniques to discover the textual patterns existed in the source code~\cite{hindle2012naturalness,gupta2017deepfix,pu2016sk_p}; following approaches proposed to learn the syntactic program embedding from the Abstract Syntax Trees (AST)~\cite{maddison2014structured,bielik2016phog,mou2016convolutional}. Such approaches addressed code completion, but fell short of more complex tasks such as program synthesis or repair, where thorough understanding and precise representations of program semantics are required. A number of new deep-learning architectures developed to specifically address this issue~\cite{allamanis2017learning,alon2018code2vec,wang2017dynamic,henkel2018code,DeFreez} tried a variety of improvements, in terms of features (static features based on program source, dynamic featured from the execution of the program~\cite{
wang2017dynamic}, abstract features from symbolic program traces~\cite{
henkel2018code}, features from the graph representation of a program~\cite{allamanis2017learning}). This diversity also makes it hard to evaluate these approaches as they do not uniformly or universally improve on existing work, but rather exhibit trade-offs across the natural characteristics of programs.

We propose a new benchmarking framework, called \coset, that aims to provide a consistent baseline for evaluating the performance of any neural program embedding. \coset captures both the human-induced variations in software (due to, e.g., coding style, algorithmic choices, code layout and structure) and algorithmic transformations to the software (due to, e.g., software refactoring, code optimization). \coset consists of a dataset of almost eighty-five thousand programs, developed by a large number of users, solving one of ten coding problems, and labeled with a number of characteristics of interest (\eg running time, pre- and post-condition, loop invariant). This dataset is paired with a number of program transformations, which when applied to any program in the dataset can produce variations of that program. These two elements (the dataset and the transformations) gives \coset the power to distinguish between various neural program embeddings with high precision.

One use of \coset is to \emph{provide a classification task} for the neural program embeddings under evaluation. The benchmark will measure the accuracy and the stability of the embedding, giving the user a precise answer on how various embeddings compare. A second use of \coset is as a \emph{debugging tool}: once a user determines that a neural program embedding fails short of some goal, the \coset program transformations can be used (in delta debugging fashion) to identify the characteristic of the programming language, the program code, or the program execution that causes the accuracy drop. 


We have conducted a pilot study with this benchmarking framework on four of the latest deep learning models. We selected \dypro~\cite{wang2017dynamic}, TreeLSTM~\cite{TreeLSTM}, the gated graph neural network (GGNN)~\cite{allamanis2017learning}, and the AST-Path neural network (APNN)~\cite{alon2018code2vec} as evaluation subjects for \coset.
Through our comprehensive evaluation, we find that \dypro achieves the best result in the semantics prediction task and is significantly more stable with its prediction results than static models, which have difficulties in capturing the program semantics. As a result, static models are much less stable against the syntax variations even when the semantics are preserved. On the other hand, the generalization from static program features as done in GGNN and APNN, while insufficiently accurate, is scalable to large or long-running programs, which overwhelm the dynamic model of \dypro. Through careful use of \coset's debugging features, we then identify a number of specific shortcomings in the tested models, from lack of support for variable types, to confusion about logging and other ancillary program aspects, and to limitations in representing APIs.

We make the following contributions: 
\begin{itemize}
	\item We design \coset, a novel benchmark framework that expresses both human and algorithmic variations in software artifacts.
	\item We propose \coset for evaluating how precise the models can 
	learn to interpret the program semantics and for identifying specific program characteristics that are the source of misclassification.
	\item We present our evaluation results including the strength and weakness 
	of each model, followed by an in-depth discussion of our findings.		
\end{itemize}


\section{Motivation and Running Example}
\label{sec:example}

\begin{wrapfigure}{R}{0.46\textwidth}
   \lstset{style=mystyle,emptylines=1,numbers=left,numberstyle=\tiny\sffamily,numbersep=10pt,aboveskip=0pt}
   \parbox{0.8\textwidth}{
\begin{lstlisting}
public static int Difference(int[] a)
{
   int swappbit=1;
   while(swappbit != 0)
   {
      swappbit=0;
      for(int i=0;i < a.Length - 1;i++)
      {
         int temp1, temp2;
         temp1=a[i];
         temp2=a[i + 1];
         if(temp1 < temp2)
         {
            a[i]=temp2;
            a[i + 1]=temp1;
            swappbit=1;
         }
      }
   }
   int b=a[0] - a[a.Length - 1];
   return b;
}
\end{lstlisting}}
   \caption{Sample code from \coset dataset, sorting an array using a Bubblesort-style strategy and then computing the difference between the smallest and the largest elements.}
   \label{fig:running}
\end{wrapfigure}

For an intuition on why programs are difficult to learn, consider the program of \autoref{fig:running}. The method \lstinline[style=mystyle,basicstyle=\small\ttfamily\bfseries]{Program.Difference()} takes as input an array, sorts it, and returns the difference between the smallest value and the largest value.

To learn the semantics of this program sufficiently well as to classify it correctly as a sorting routine using the Bubblesort strategy requires bridging the gap between the syntactic representation and its runtime execution. This gap is typically expressed in abstract interpretation by considering a program to be the sum-total of all its possible execution traces and considering a program's source code to be an abstraction of the program-as-traces model. Intuitively this distinction implies that learning to classify programs may be limited in accuracy when using only the source code, and correspondingly benchmarks should focus on samples whose source code and execution traces are not trivially related.

The gap between syntax and (runtime) semantics is characterized by several properties, which become the requirements for our \coset benchmark. In other words, we wish for \coset to have enough training and test samples to cover the whole range of distinctions between syntax and semantics.

First, the gap between program syntax and semantics is manifested in that similar programs of similar syntax may have to vastly different semantics. For example, consider the two sorting implementations, both sorting an array via two nested loops, both comparing the current element to its successor, and both swapping them if the order is incorrect. Yet the two functions could implement different sorting strategies, namely Bubblesort and Insertionsort. Therefore minor syntactic discrepancies can lead to significant semantic differences. Our \emph{first requirement} for the benchmark is to \emph{include sufficient samples that vary both syntactically and semantically, though solving the same software task.}

Second, program statements are almost never interpreted in the order in which the corresponding token sequence appears in the source code (the only exception being straight-line programs, i.e., ones without any control-flow statements). For example, a conditional statement only executes one branch each time, but its token sequence is expressed sequentially as multiple branch bodies. Similarly, when iterating over a looping structure at runtime, it is unclear in which order any two tokens are executed when considering different loop iterations. In \autoref{fig:running} the statement on line~23 may or may not execute after the statement on line~25 from a previous loop iteration. A related characteristic is that the order of independent statements (e.g., lines~19 and~20) is arbitrary and the learning model needs to be evaluated for syntactic-ordering biases. The \emph{second requirement} for our benchmark is to \emph{include samples that vary only in the syntactic order of statements, without changing semantics.}

Third, many tokens in the source code are not relevant to program semantics. For example, variable names do not affect the results computed by a program, but rather the dependencies between variables (both data and control dependencies) play an essential role in defining program semantics. In \autoref{fig:running} the variables \lstinline[style=mystyle,basicstyle=\small\ttfamily\bfseries]{swappbit}, \lstinline[style=mystyle,basicstyle=\small\ttfamily\bfseries]{i}, \lstinline[style=mystyle,basicstyle=\small\ttfamily\bfseries]{temp1}, \lstinline[style=mystyle,basicstyle=\small\ttfamily\bfseries]{temp2}, \lstinline[style=mystyle,basicstyle=\small\ttfamily\bfseries]{b}, and \lstinline[style=mystyle,basicstyle=\small\ttfamily\bfseries]{a} could have any other names, distinct from each other, and the program would behave the same way. The \emph{third requirement} for our benchmark is to \emph{include samples with arbitrarily values for syntactic tokens that do not affect semantics.}


\section{Benchmark Design}
\label{sec:benchmark}

The requirements from the previous section determine the need for a wide range and wide diversity of programs, both in terms of source code and in terms of execution behavior. Before describing \coset's design, we add an additional requirement, that of \emph{non-adversarial, natural style} for our dataset samples. The goal of \coset is to provide the capability of testing learning models on their accuracy in capturing program semantics, and we wish to explicitly exclude adversarial examples from this benchmark. The rationale is that adversarial examples, which underwent program obfuscation in order to make their analysis and understanding exceedingly hard, are an orthogonal problem that requires separate solutions and accordingly a distinct benchmark. For example, a common obfuscation used by malicious programs is that of embedding an interpreter. The original (malicious) program is replaced by a new program that consists of an interpreter and an reimplementation of the original program in a randomly generated language, to be handled by the interpreter. Understanding any adversarial examples obfuscated this way requires addressing three tasks: (1) identifying the presence of an interpreter, (2) learning the semantics of that interpreter, and (3) learning the semantics of the source code embedded alongside of the interpreter. \coset as a benchmark focuses only on task (3) and thus explicitly excludes adversarially generated samples.

With this additional requirement in mind, we design the benchmark to include samples that capture normal variation in terms of syntax, style, algorithms, etc. arising in day-to-day programming work. This means using program samples from a diverse set of programmers, all solving the same coding problem, as well as including tools to transform these samples according to two common steps in the software-development lifecycle, refactoring and optimization.


\subsection{Program Dataset}
The dataset consists of 84,165 programs in total. They are obtained from a popular online coding platform. Programs were written in several different languages: Java, C\# and Python. All programs solve a particular coding problem. We hand picked the problems to ensure the diversity of the programs in the dataset. Specifically it contains introductory programming exercises for beginners, coding puzzles that exhibit considerable algorithmic complexity and challenging problems frequently appearing on coding interviews. We split the whole dataset into a training set containing 64,165 programs, a validation set of 10,000 programs, and a test set of the remaining 10,000 programs.


\subsection{Program Labels}

The dataset have been manually analyzed and labeled on the basis of operational semantics (\eg Bubblesort, Insertionsort, Mergesort, \etc, for a sorting routine). We allow certain kind of variations to keep the total number of labels manageable. For example, we ignore local variables allocated for temporary storage, the iterative style of looping or recursion, sort order: descending or ascending, \etc

The work is done by fourteen PhD students and exchange scholars at University of California, Davis. They come from different research backgrounds such as programming language, database, security, graphics, machine learning, etc. All of them have been interviewed and tested for their knowledge on program semantics. To reduce the labeling error, we distributed programs solving the same coding problem mostly to a single person and have them cross check the results for validation. The whole process took more than three months to complete. In the end, we defined 38 labels in total, with more than 2,000 programs on average for each label.\footnote{Readers are encouraged to consult the supplemental material.}


\subsection{Program Transformations}
\label{sec:pt}


In this section, we introduce the transformations we apply to \coset for generating program variants. They serve two purposes in our experiments: debugging the model and gauging the model stability. The former refers to identifying the reason of misclassification; the latter measures how stable are the models against software variations, both of which will be explained in details in \autoref{sec:exp}. \coset considers three types of transformations, semantics-preserving, semantics-approximating, and semantics-changing.

\paragraph{Semantics-Preserving Transformations:}
This category can be further split into compiler
optimizations (\ie for improving the performance of
a program in code compilation) and software
refactoring (\ie for improving the code readability and
maintainability in software development). Even though
they are designed for different settings and purposes,
all of them preserve the semantics of the original programs.
In this paper we choose Constant and Variable Propagation (CVP),
Dead Code Elimination (DCE), Loop Unrolling (LU) and Hoisting
for compiler optimization and Variable Renaming (VR), Nested
Condition Simplification (NCS), Control Flag Removal (CFR) and
Control Statement Unification (CSU) for software refactoring.
Interested readers are invited to consult Aho et al.'s compiler textbook~\cite{Aho:2006} and the Eclipse IDE documentation~\cite{ecl,goth2005beware} for the examples of each transformation.

\paragraph{Semantics-Approximating Transformations:}
Semantics-approximating transformations change a program in minor ways producing a new program with semantics \emph{close} to that of the original. In particular we adopt two kinds of transformations.
First we change type names to others within the same group.
For example, we replace \texttt{int} with \texttt{long} in
the group of integer types, or \texttt{double} with
\texttt{float} in the group of floating-point types. Second
we also change an API in the program to another of the similar semantics
such as changing \texttt{Array.Sort(a:array)} to
\texttt{Array.Sort(a:array,c:comparer)} from Microsoft's .NET Framework for C\#.

\paragraph{Semantics-Changing Transformations:}
We also include transformations that change the
semantics of a program. Those transformations are
necessary solely for the purpose of debugging a model.
Specifically \coset can remove error handling code from a
program and can fix a buggy program with SARFGEN~\cite{WangPLDI},
a productized repair tool often used in Massive Open Online
Courses (MOOCs).



\section{Evaluation}
\label{sec:exp}

\lstdefinestyle{I}{style=mystyle,basicstyle=\small\ttfamily\bfseries}

In this section, we describe in details how we use \coset to evaluate the strengths and weaknesses of each deep neural architecture by measuring two characteristics---accuracy and stability.

\subsection{Evaluating Classification Accuracy Using \coset}
\label{subsubsec:acc}

We train GGNN, APNN, TreeLSTM and \dypro on \coset 
to predict the semantic label of a test program. In 
this experiment we use prediction accuracy and F1 score 
to measure classification results.

\begin{figure*}[t!]
	\begin{subfigure}[b]{0.375\textwidth}
		\begin{center}
			\centerline{\includegraphics[width=\columnwidth]{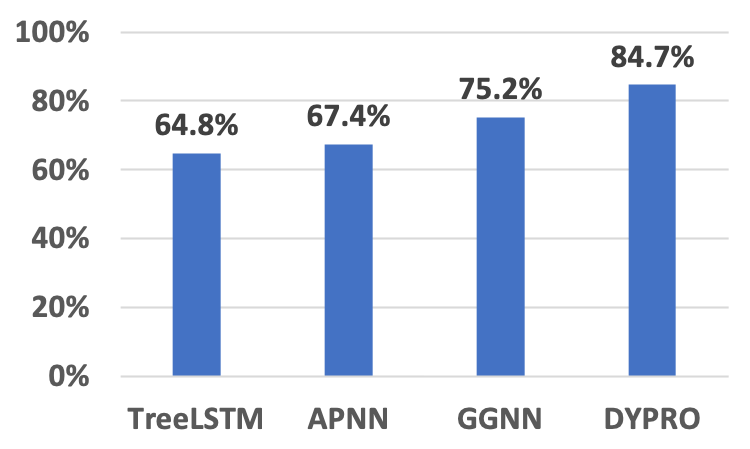}}
			\caption{Prediction (accuracy).}
			\label{fig:acc}
		\end{center}
	\end{subfigure}
	\begin{subfigure}[b]{0.375\textwidth}		
		\begin{center}
			\centerline{\includegraphics[width=\columnwidth]{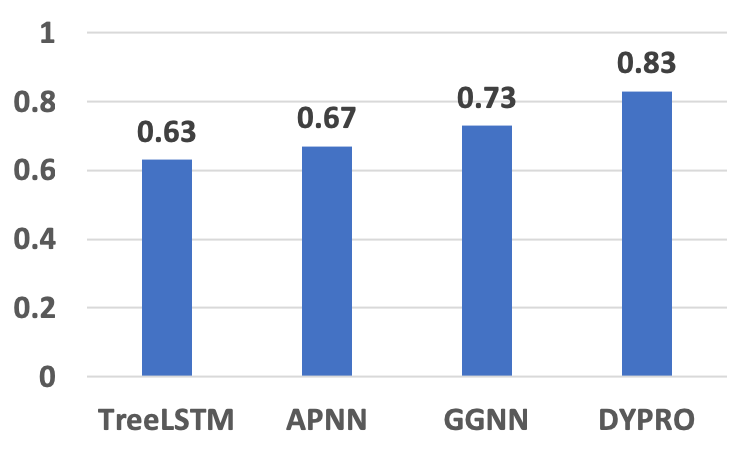}}
			\caption{Prediction (F1 score).}
			\label{fig:accF1}
		\end{center}
	\end{subfigure}
	\begin{subfigure}[b]{0.375\textwidth}		
		\begin{center}
			\centerline{\includegraphics[width=\columnwidth]{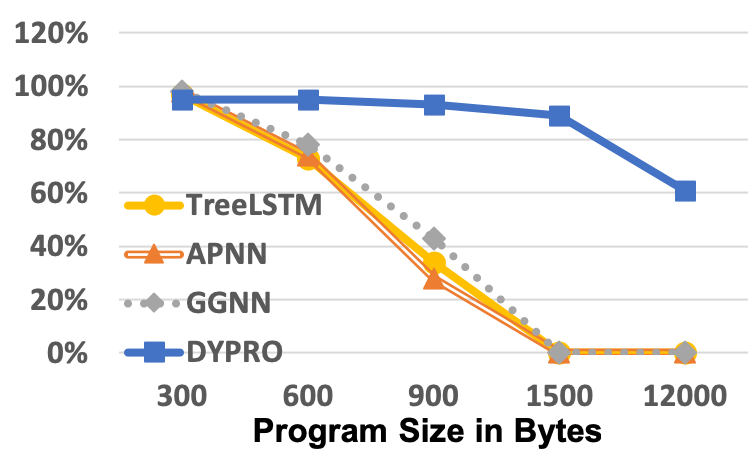}}
			\caption{Scalability (accuracy).}
			\label{fig:scas}
		\end{center}
	\end{subfigure}
	\begin{subfigure}[b]{0.375\textwidth}		
		\begin{center}
			\centerline{\includegraphics[width=\columnwidth]{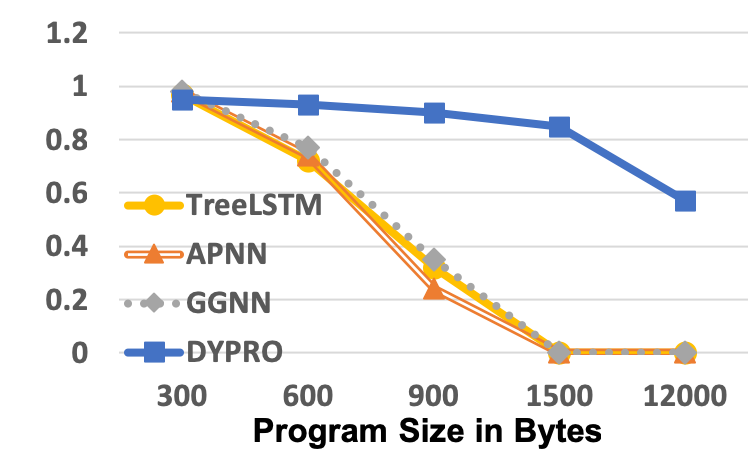}}
			\caption{Scalability (F1 score).}
			\label{fig:scasF1}
		\end{center}
	\end{subfigure}
	
	\caption{Evaluation Results.}
	\label{fig:ss}
	\vskip -0.1in
\end{figure*}

\paragraph{Results.}
\autoref{fig:acc} depicts the prediction accuracy of all models on \coset. Overall, \dypro comes on top as the most accurate model at 84.7\%, with GGNN achieves 75.2\% prediction accuracy. In terms of F1 score (\autoref{fig:accF1}), \dypro and GGNN keep their top positions while APNN and TreeLSTM swap their places at the bottom of the ranking. The other metric of interest is scalability in terms of program size. In order to compare GGNN, APNN, and TreeLSTM (which operate on source code, typically measured as number of lines) on one hand, and \dypro (which operates on execution trace, measured as number of state changes) on the other, we unify the program-size dimension as number of bytes, with a line of code being on average 15~bytes in length and a trace state changes 20~bytes. As shown in Figures~\ref{fig:scas} and~\ref{fig:scasF1}, accuracy quickly drops as program size increases, with static models (GGNN, APNN, and TreeLSTM) losing all accuracy once a program exceeds 100 lines.

\begin{figure*}
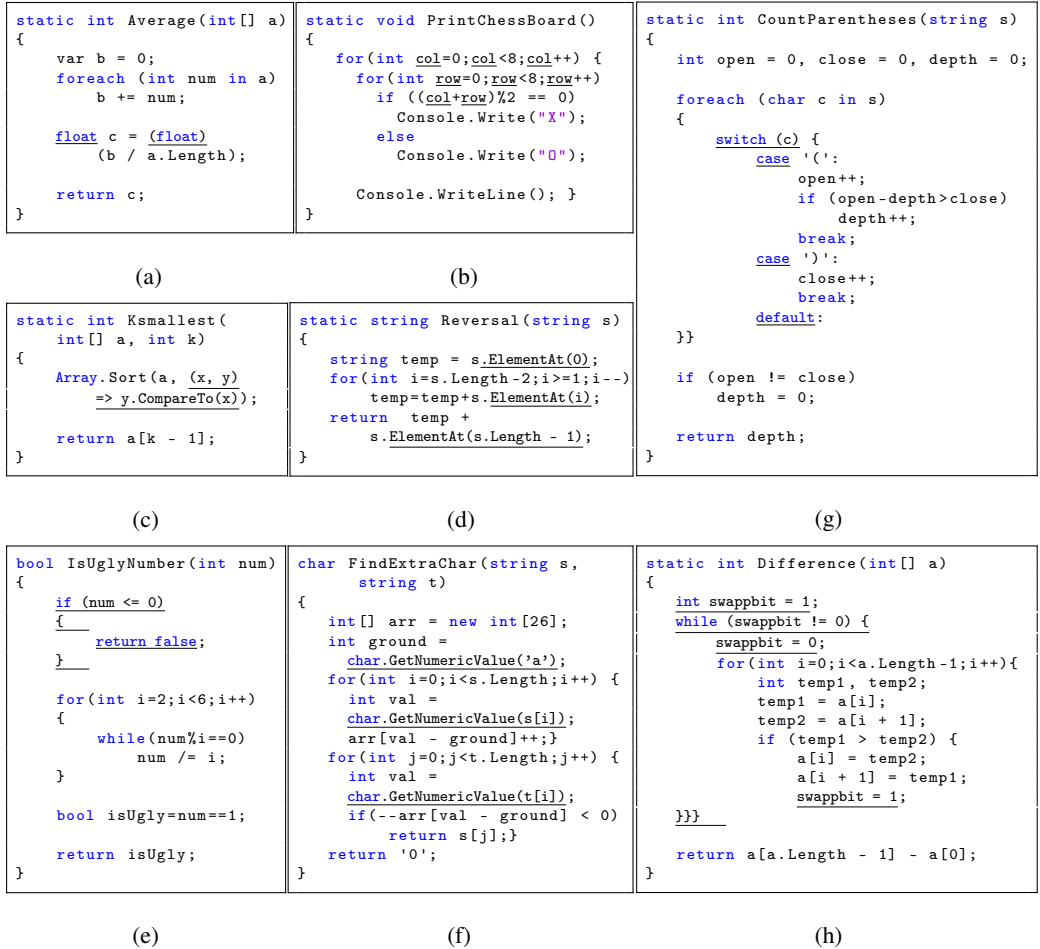

	\begin{subfigure}[t]{0.58\textwidth}
	\vskip 0pt
		\begin{subfigure}{0.44\textwidth}
			\lstset{style=mystyle}
\begin{lstlisting}[linewidth=3.59cm]
static int Average(int[] a)
{
    var b = 0;
    foreach (int num in a)
        b += num;

    @\underline{\textcolor{blue}{float}}@ c = @\underline{(\textcolor{blue}{float})}@
        (b / a.Length);

    return c;
}
\end{lstlisting}		
			\caption{}
			\label{fig:a}
		\end{subfigure}
		\begin{subfigure}{0.525\textwidth}	
			\lstset{style=mystyle}
\begin{lstlisting}
static void PrintChessBoard()    
{
   for(int @\underline{col}@=0;@\underline{col}@<8;@\underline{col}@++) {    
     for(int @\underline{row}@=0;@\underline{row}@<8;@\underline{row}@++)
       if ((@\underline{col}@+@\underline{row}@)%2 == 0)
         Console.Write("X");
       else
         Console.Write("O");

     Console.WriteLine(); }
}
\end{lstlisting}
			\caption{}
			\label{fig:b}
		\end{subfigure}
		\begin{subfigure}{0.43\textwidth}
			\lstset{style=mystyle}
\begin{lstlisting}[linewidth=3.51cm,basicstyle=\linespread{.67}\fontsize{5.6}{10.8}\ttfamily\bfseries]
static int Ksmallest(
    int[] a, int k)
{
    @\textcolor{blue}{Array}@.Sort(a, @\underline{(x, y)}@ 
        @\underline{=> y.CompareTo(x)}@);

    return a[k - 1];
}
\end{lstlisting}		
			\caption{}
		    \label{fig:c}
		\end{subfigure}
		\begin{subfigure}{0.535\textwidth}	
			\lstset{style=mystyle}
\begin{lstlisting}[basicstyle=\linespread{.68}\fontsize{5.6}{10.8}\ttfamily\bfseries]
static string Reversal(string s)
{
   string temp = s@\underline{.ElementAt(0)}@;
   for(int i=s.Length-2;i>=1;i--)
       temp=temp+s.@\underline{ElementAt(i)}@;
   return  temp + 
       s.@\underline{ElementAt(s.Length - 1)}@;
}
\end{lstlisting}
			\caption{}
			\label{fig:d}
		\end{subfigure}	
		\begin{subfigure}{0.43\textwidth}
			\lstset{style=mystyle}
\begin{lstlisting}
bool IsUglyNumber(int num)
{
    @\underline{\textcolor{blue}{if} (num <= 0)}@
    @\underline{\{\textcolor{white}{11}                    }@
        @\underline{\textcolor{blue}{return false}}@;
    @\underline{\}\textcolor{white}{11}                    }@

    for(int i=2;i<6;i++)
    {
        while(num%i==0)
            num /= i;
    }

    bool isUgly=num==1;

    return isUgly;
}
\end{lstlisting}		
			\caption{}
			\label{fig:e}
		\end{subfigure}
	    \hspace{.09cm}
		\begin{subfigure}{0.535\textwidth}	
			\lstset{style=mystyle}
\begin{lstlisting}[upquote=true,linewidth=4.36cm]
char FindExtraChar(string s, 
      string t)
{
   int[] arr = new int[26];
   int ground =
     @\underline{\textcolor{blue}{char}.GetNumericValue('a')}@;
   for(int i=0;i<s.Length;i++) { 
     int val = 
     @\underline{\textcolor{blue}{char}.GetNumericValue(s[i])}@;
     arr[val - ground]++;}
   for(int j=0;j<t.Length;j++) {
     int val =
     @\underline{\textcolor{blue}{char}.GetNumericValue(t[i])}@;
     if(--arr[val - ground] < 0)
         return s[j];}
   return '0';
}
\end{lstlisting}
			\caption{}
			\label{fig:f}
		\end{subfigure}	
	\end{subfigure}
\;
	\begin{subfigure}[t]{0.35\textwidth}		
	\vskip 0pt
		\begin{subfigure}{\textwidth}		
			\lstset{style=mystyle}
\begin{lstlisting}[basicstyle=\linespread{.696}\fontsize{6}{10.8}\ttfamily\bfseries,upquote=true,linewidth=5.1cm]
static int CountParentheses(string s)
{
   int open = 0, close = 0, depth = 0;

   foreach (char c in s)
   { 
       @\underline{\textcolor{blue}{switch} (c)}@ { 
           @\underline{\textcolor{blue}{case}}@ '(':
               open++;
               if (open-depth>close)
                   depth++;
               break;
           @\underline{\textcolor{blue}{case}}@ ')':
               close++;
               break;
           @\underline{\textcolor{blue}{default}}@:
   }}

   if (open != close)
       depth = 0;

   return depth;
}
\end{lstlisting}
			\caption{}
			\label{fig:g}			
		\end{subfigure}	
		\\	
		\begin{subfigure}{\textwidth}		
			\lstset{style=mystyle}
\begin{lstlisting}[basicstyle=\linespread{.666}\fontsize{6}{10.8}\ttfamily\bfseries,linewidth=5.1cm]
static int Difference(int[] a)
{
   @\underline{\textcolor{blue}{int} swappbit = 1}@;
   @\underline{\textcolor{blue}{while} (swappbit != 0) \{}@ 
       @\underline{swappbit = 0}@;
       for(int i=0;i<a.Length-1;i++){ 
           int temp1, temp2;
           temp1 = a[i];
           temp2 = a[i + 1];
           if (temp1 > temp2) { 
               a[i] = temp2;
               a[i + 1] = temp1;
               @\underline{swappbit = 1}@;
   @\underline{\}\}\}\textcolor{white}{111}}@

   return a[a.Length - 1] - a[0];
}
\end{lstlisting}
			\caption{}
			\label{fig:h}			
		\end{subfigure}				
	\end{subfigure}
		
	\caption{Misclassifications triggered by \coset. Underlined code is the root cause of the misclassification, as identified by \coset's debugging capability. See \autoref{sec:exp} for a detailed analysis.}
	\label{fig:mis}
	\vskip -0.1in
\end{figure*}

\paragraph{Discussion of Misclassification Types.}
\coset is able to trigger many types of misclassifications. To gain a better 
understanding of the errors each model made, 
we have conducted a large scale manual inspection 
and summarize our findings below.

\emph{Types and Variable Names:}
Type name can cause issues. For example, \autoref{fig:a} 
shows a program in \coset that 
causes APNN to misclassify. 
Replacing \lstinline[style=I]{float} to \lstinline[style=I]{double} fixes this error. 
As a more subtle example, program in \autoref{fig:b} is misclassified by all of the static models. The error is due to variable 
\texttt{col} and \texttt{row}. Since vast majority of programs 
in \coset use \texttt{row} (\resp \texttt{col}) as the induction 
variable in the outer (\resp inner) loop
, static models do not recognize the same semantics a 
program denotes after some of its variables (\ie \texttt{row} and \texttt{col}) 
are interchanged. 
It suffices to conclude 
that static models weigh variable names in their prediction, 
where in fact they should not be considered for any semantics-based classification task.

\emph{Syntactic Structure:}
To capture human programming habits in terms of code style and code structure, 
\coset exposes a uneven distribution of code samples in terms 
of program syntax (\ie some are more frequently used 
than others), which helps to trigger misclassifications. 
For example, the \lstinline[style=I]{switch} statement 
in program~\ref{fig:g} is the reason of the error. 
In other words, models would have correctly classified 
the program if the \lstinline[style=I]{switch} is converted 
to a \lstinline[style=I]{if-else}, a far more popular selection 
statement. 

\emph{Scalability:}
\coset also reveals scalability being a major issue  
among all models including both static and dynamic. 
As the size of programs/execution traces increases, 
all models suffer notable drops in accuracy (\resp 
F1 score) depicted in \autoref{fig:scas} (\resp 
\autoref{fig:scasF1}). In particular APNN is shown 
to be the most vulnerable.

\emph{API Usage:}
\coset discloses issues among static models in 
generalizing API calls. Figures~\ref{fig:c} and~\ref{fig:d} 
show two examples. \autoref{fig:c} (\resp \autoref{fig:d}) 
can be fixed by replacing the underlined API with 
\texttt{\textcolor{blue}{Array}.Sort(a)} (\resp indexing operator []). Granted, 
\coset bears some blame as it contains very few programs 
of the same label as the two examples that has the exact identical 
API signatures, but to expect a sufficient coverage of all 
API prototypes (\eg 17 in total for \texttt{\textcolor{blue}{Array}.Sort} according to 
Microsoft .NET framework 4.7) is a challenging task for any 
dataset. Therefore generalizing to largely similar despite 
partially unseen API is necessary and immediate. In addition, 
in both cases such API variations account for a 
small portion of the whole program. In particular we find 86.2\% (\resp 68.9\%) 
of the programs in the training set have a very similar body 
(\ie differ by less than eight tokens) to \autoref{fig:c} (\resp \autoref{fig:d}) 
according to DECKARD~\cite{Jiang}, arguably 
the state-of-the-art clone detection algorithm. As a result, it is reasonable to expect static 
models to be immune from the minor API changes.

\emph{Error Handling:}
\coset shows error handling code can be another 
cause of misclassification. \autoref{fig:e} depicts 
an example that is misclassified by APNN and TreeLSTM. 
By removing the \lstinline[style=I]{if} clause at the beginning of the 
function, both models produce the correct result. This is 
another example \coset demonstrates that static models 
are unstable against syntax change that barely affects the 
program semantics.

\emph{Buggy Code:}
\coset reveals the \dypro model as the most susceptible 
to buggy programs. The reason being the 
dynamic nature of program representation \dypro 
adopts. That is if there is a bug in the program, its runtime dynamic 
traces will likely lead to greater discrepancies than 
the syntactic representations (\eg tokens, ASTs, \etc). 
Take the program in \autoref{fig:f} for example, 
the bug lies in the API call in \lstinline[style=I]{int val = char.GetNumericValue(`a')} 
which converts a character of digit to an integer. In 
contrast what the programmer intended is to directly 
get the ascii code/unicode of the character using \lstinline[style=I]{int val = `a'}. 
Unfortunately this bug leads to a chaotic dynamic 
representation (\ie \lstinline[style=I]{char.GetNumericValue()} will be evaluated to 
-1 for any non-digit character), \dypro is not able to 
overcome. In comparison, all static models display a stronger fault tolerance by correctly classifying the buggy program. 


\emph{Semantic Property:}
\coset gives insights into the limitations of 
static models in learning semantic program properties. 
The program in \autoref{fig:h} triggers a misclassification 
of this kind on all static models. Worth mentioning the program 
does not implement a standard Bubblesort strategy, in which 
the inner loop systematically shifts its bound towards the beginning 
of an array after bigger numbers are being bubbled up to the end of 
the array. Instead the inner loop in the program keeps sorting the 
entire array until reaching an iteration where no items are swapped, 
at which point the array is sorted. 

Our methodology for analyzing this misclassification starts with formally defining the semantic label $\mathit{L}$ of \textit{Bubblesort} to apply to programs that have a function satisfying the following properties:
\begin{enumerate}
   \item time complexity of $\mathit{O}(n^{2})$;
   \item input type signature of $a:array$ and post-condition that the array $a$ is sorted, $\forall i\in[0,a.Length-1).\; a[i] \leq a[i+1]$;
   \item having a nested loop structure and an invariant of the outer loop assuming $i$ is the number of its iterations, $\forall j\in[a.Length-i,a.Length-1].\; a[j-1] \leq a[j]$.
\end{enumerate} 
Similarly we give definitions of all other labels in \coset.
To identify the source of the misclassification (\ie a program not being classified as $L = \textit{Bubblesort}$), we repeat the classification task for a broader set of classes, using one property at a time as a new label definition. For example, using property (1) alone we collect all programs with time complexity $\mathit{O}(n^{2})$ in \coset as the training data for a label $L'$. Similarly for property (2) we use all programs that implement a sorting routine and label them $L''$; and for (3) we collect all programs that display the bubbling behavior under label $L'''$. Results show static models incorrectly predict the program in \autoref{fig:h} only when using property (3) as label definition. In other words, the static models GGNN, APNN, and TreeLSTM fail to learn a sufficiently expressive discriminator to understand the non-standard bubbling sort strategy. This methodology demonstrates \coset's utility in discovering the limitations of models in learning semantic properties.

\begin{table}
	\begin{center}
			\begin{tabular}[t]{crrrrrrrr} 
				\toprule
				\multirow{2}{*}{Models}
				
				& \multicolumn{4}{c}{Compiler Optimization}     
				& \multicolumn{4}{c}{Software Refactoring}    \\ 
				\cmidrule{2-9}
				& \multicolumn{1}{c}{CVP}
				& \multicolumn{1}{c}{DCE}
				& \multicolumn{1}{c}{LU}
			    & \multicolumn{1}{c}{Hoisting}
   				& \multicolumn{1}{c}{VR}
				& \multicolumn{1}{c}{NCS}
				& \multicolumn{1}{c}{CFR}
				& \multicolumn{1}{c}{CSU}
			    \\				

                \midrule
				GGNN  &4.2\% &5.6\% &19.1\% &7.8\% &8.4\% &13.5\%  &11.0\% &17.2\% \\										
				APNN  &7.4\% &12.7\%  &27.7\% &11.7\% &7.1\%  &16.3\% &14.9\% &22.8\% \\										
				TreeLSTM &7.2\% &5.5\% &14.6\% &6.4\% &5.4\% &12.8\% &8.5\% &19.7\%\\

				\dypro &2.6\% &4.5\% &5.3\% &1.7\% &0.0\% &0.0\% &0.0\% &0.0\%\\

				\bottomrule
			\end{tabular}
		\caption{Results of measuring the stability using \coset, as percentage of changed predictions when applying semantics-preserving transformations.}
		\label{Table:rob}		
	\end{center}	
\end{table}

\subsection{Evaluating Model Stability Using \coset}
\label{subsubsec:rob}

In this experiment we evaluate how stable the models are with their predictions. The reason stability is a relevant and important metric is because changes in source code are inevitable, and a model that is resilient against changes, especially the semantics-preserving kind, will yield many benefits. 
As mentioned in the benchmark design (\autoref{sec:benchmark}), \coset applies normal, standard, and widely adopted program transformations to simulate how software evolves during its normal lifecycle (\eg code optimization or software refactoring).

Given the models that are trained for the experiment in \autoref{subsubsec:acc} 
and \coset's testing program, we apply code transformations to generate a new test set, preserving semantics for all samples. Programs with no applicable transformations are excluded from the new test set. We then examine if models hold on to their prior predictions.
\autoref{Table:rob} depicts the results. The number in each cell denotes 
the percentage of programs in the new test set on which a model \emph{changed} its prediction. 
Overall, all models display decent stability against software transformations. No transformation was able to sway 30\% or more predictions for any model. For the static models, TreeLSTM is shown to be the most stable deep neural architecture while APNN is 
the most sensitive model to the transformations. Unsurprisingly, \dypro handles almost all semantics-preserving transformations as its features are extracted from execution traces and thus largely unaffected by these source-code transformations.

\section{Related Work}
\label{sec:related}



MNIST~\cite{lecun1998gradient} (Modified National Institute of Standards and Technology database) is a database of handwritten digits that is widely used for training and testing machine learning algorithms. The MNIST database contains 60,000 training images and 10,000 testing images. There have been many attempts to achieve the lowest error rate, among which some indeed have accomplished "near-human performance".
ImageNet~\cite{imagenet_cvpr09} is a visual database designed for visual object recognition research. More than 14 million images have been hand-annotated to indicate what objects are pictured. ImageNet contains more than 20,000 categories, each of which consists of several hundred images. Since 2010, ImageNet has been adopted in an annual contest, the ImageNet Large Scale Visual Recognition Challenge (ILSVRC), where machine learning models compete to correctly classify and detect objects and scenes.
CIFAR-10~\cite{krizhevsky2009learning} (Canadian Institute For Advanced Research) is another database of images that are commonly used to train machine learning and computer vision algorithms. It is one of the most widely used datasets for machine learning research. The CIFAR-10 dataset contains 60,000 32x32 color images in 10 different classes \eg cars, birds, cats, \etc
Labeled Faces in the Wild~\cite{huang2008labeled} is a database of face photographs designed for studying the problem of unconstrained face recognition. The dataset contains more than 13,000 images of faces collected from the web. Each face has been labeled with the name of the person pictured. 1680 of the people pictured have two or more distinct photos in the data set. The only constraint on these faces is that they were detected by the Viola-Jones face detector.

Unlike the existing databases designed almost exclusively for training and testing machine learning models in computer vision and image processing domain, we present \coset to evaluate how precise deep neural architecture can learn the semantics of a program. In addition we also introduce a suite of program transformations for assessing the stability of model prediction and debugging the classification mistake.

\section{Conclusion}
\label{sec:con}


We introduce a benchmark framework called \coset for evaluating the accuracy and stability of neural program embeddings and related neural network architectures proposed for software classification task. \coset consists of a set of natural, non-adversarial source-code samples from a variety of programmers tasked with well-defined coding challenges, and supplements this set with source-code transformations representative of common software optimization and refactoring. In our evaluation we observed that \coset was effective in measuring differences among various models proposed in literature and provided debugging capabilities to identify the root causes of misclassifications.


\bibliography{references}

\end{document}